\documentclass[journal]{IEEEtran}

\usepackage{amsmath,amssymb}
\usepackage{booktabs}
\usepackage{array}
\usepackage{graphicx}
\usepackage{xcolor}
\usepackage{microtype}
\usepackage{multirow}
\usepackage{siunitx}
\usepackage{url}
\usepackage{balance}
\usepackage{placeins}
\usepackage[hidelinks]{hyperref}

\definecolor{tglblue}{HTML}{315D83}
\definecolor{tglteal}{HTML}{2F7D78}
\definecolor{tglamber}{HTML}{C98A32}
\definecolor{tglcoral}{HTML}{BE5B54}
\definecolor{tglgray}{HTML}{55616D}

\hypersetup{
  pdftitle={Teach and Grow: An Agent-Centered Architecture for General Robot Learning},
  pdfauthor={Chang Nie, Zhe Liu, and Hesheng Wang},
  pdfsubject={Agent-centered robot manipulation and skill learning; an implementation with GPT-6 Astra and Codex},
  pdfkeywords={AI agents, GPT-6 Astra, Codex, robot manipulation, Teach-and-Grow Learning, few-shot robot learning, skill composition}
}

\graphicspath{{figures_imagegen_v9/}{figures/}{figures_ai/}}

\title{Teach and Grow: An Agent-Centered Architecture for General Robot Learning}
\author{Chang Nie, Zhe Liu, and Hesheng Wang%
\thanks{Chang Nie, Zhe Liu, and Hesheng Wang are with the School of Automation and Intelligent Sensing, Shanghai Jiao Tong University, and the Shanghai Key Laboratory of Navigation and Location Based Services, Shanghai 200240, China. Chang Nie: \mbox{changniep@gmail.com}. Corresponding author: Hesheng Wang (e-mail: \mbox{wanghesheng@sjtu.edu.cn}).}}

\begin{document}
\maketitle

\begin{abstract}
Vision-language-action (VLA) and world-action models typically absorb unfamiliar manipulation tasks through additional robot data collection and policy optimization. This recurring retraining burden slows the acquisition of new behavior. We present Teach-and-Grow Learning (TGL), a training-free architecture that turns a few successful demonstrations into reusable robot skills. Task acquisition requires no gradient updates, fine-tuning, or reinforcement learning: pretrained model weights remain fixed as the robot expands its explicit knowledge. Teaching is an accelerator, not a precondition, because the agent can also drive the robot directly, and demonstrations mainly improve reliability. Our implementation uses OpenAI GPT-6 Astra for multimodal reasoning and Codex to connect the agent to robot tools. The agent identifies subgoals shared across demonstrations, expresses them as closed-loop Skill Blocks, and grounds each block in the current scene. Physical feedback guides the next action and any recovery. Verified behaviors enter a persistent Skill Library; Experience Memory records the conditions and repairs that inform later decisions. TGL reaches 99.9\% mean success on four LIBERO suites and 92.4\% on seven LIBERO-Plus perturbation categories. Controlled studies show that taught blocks persist and improve related-task execution under the same model weights and executors. We further formulate a scaling hypothesis that relates effective reusable experience to falling future-task error and teaching demand. Code and demonstration videos: \url{https://tgl.changnie.top}.
\end{abstract}


\section{Introduction}
\label{sec:introduction}

Vision-language-action (VLA) and world-action models map observations and language directly to actions, replacing task-specific systems with one scalable policy. RT-1 and RT-2 showed that large-scale learning can extend robot behavior~\cite{brohan2022rt1,zitkovich2023rt2}, and later generalist policies and world-action models widened the range of tasks and physical dynamics they cover~\cite{octoteam2024octo,kim2024openvla,black2024pi0,ye2026dreamzero}. The extrapolation is that larger models and more data will eventually yield general robot intelligence. Their reliability, however, is bounded by validated physical coverage. A capability is stored in the same shared weights as everything else, so when an unfamiliar object, sensor, body, or contact falls outside that coverage, acquiring it means collecting more robot data and optimizing those weights again. Every such gap therefore arrives as another training cycle.

\begin{figure}[!t]
    \centering
    \includegraphics[width=\columnwidth]{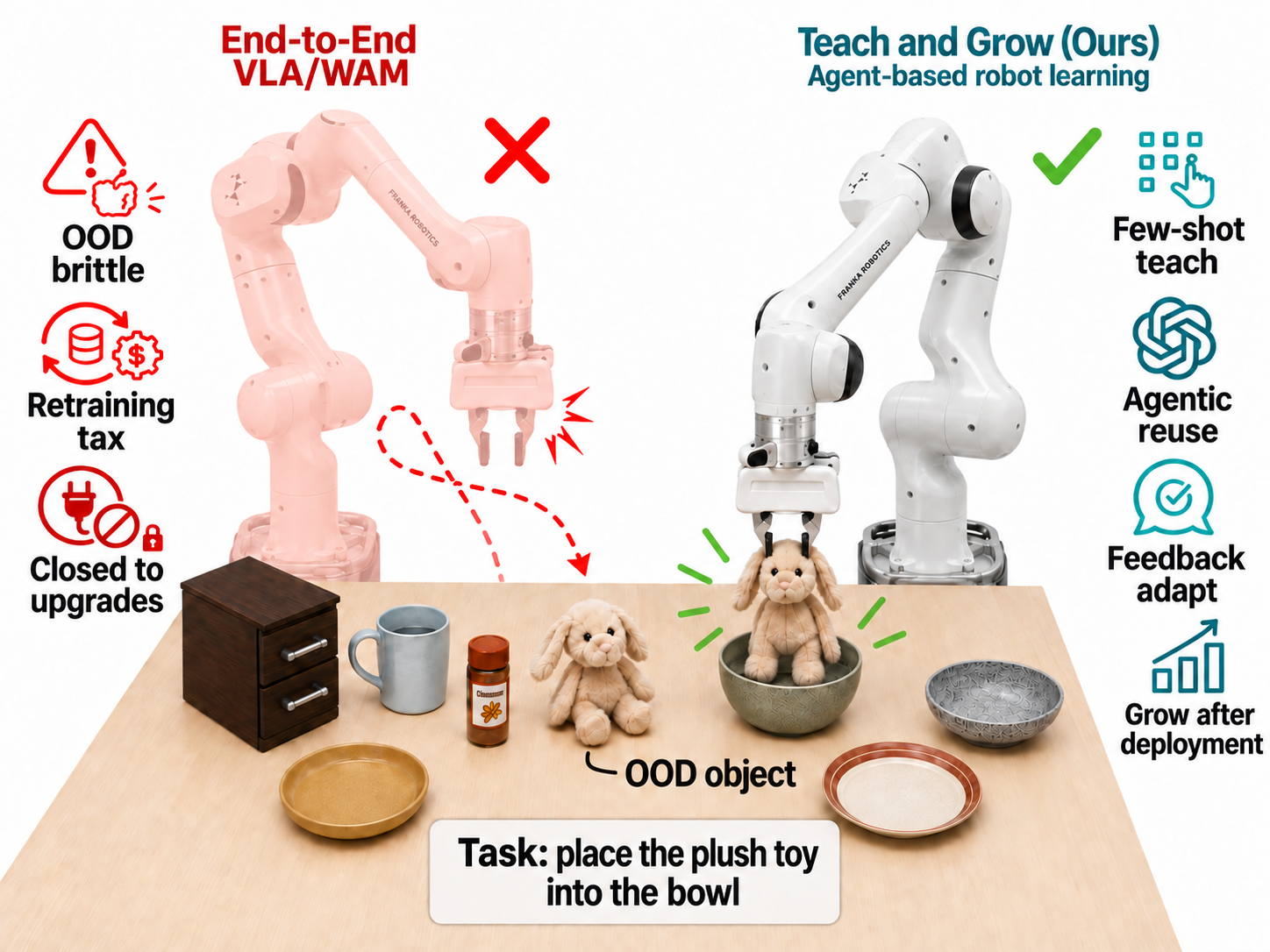}
    \caption{Two ways to acquire an unfamiliar pick-and-place behavior. \emph{Left:} the illustrated VLA/WAM route incorporates the new plush-toy interaction through a policy-training cycle. \emph{Right:} TGL uses sparse teaching to form reusable Skill Blocks without policy training, composes them through an AI agent, and checks physical outcomes before retaining the behavior for later tasks.}
    \label{fig:teaser}

\end{figure}

Such a cycle carries three costs at once: new data collection, another optimization run, and regression checking against everything the policy already supported. We call this recurring burden the \emph{retraining tax}. Its most uncomfortable property is that a local failure demands a global repair. Fixing one unfamiliar bowl does not leave behind a separately addressable ``bowl-retention'' capability that a person can inspect, test, or reuse; the change spreads through the weights, and as competence accumulates, every update has to account for more of it.

The price of that cycle is set by a resource robotics does not have. Generalization in large models is bought with data, and language models could buy it cheaply: text, code, books, and webpages had already been produced for other purposes, so pretraining drew on a corpus that was vast and ready-made. Robot learning has no such corpus. Its evidence, observations paired with actions and with their physical consequences on a particular body, does not exist until a robot or a simulator is run, which is why robot datasets are assembled through deliberate collection campaigns~\cite{openx2023,khazatsky2024droid,wu2024robomind,agibot2025world}. The generality that emerges in language models must therefore be purchased, in robotics, with physical interaction at a cost far above pretraining on text. The difficulty compounds because physical conditions multiply rather than add: object pose, camera geometry, clutter, material, and embodiment interact, so covering one factor does not cover their combinations. Web-scale pretraining helps a robot interpret an instruction or recognize an object, but it does not by itself supply the geometry, dynamics, and contact information that reliable execution requires~\cite{fang2025intact}.

\begin{figure*}[!t]
    \centering
    \includegraphics[width=0.90\textwidth]{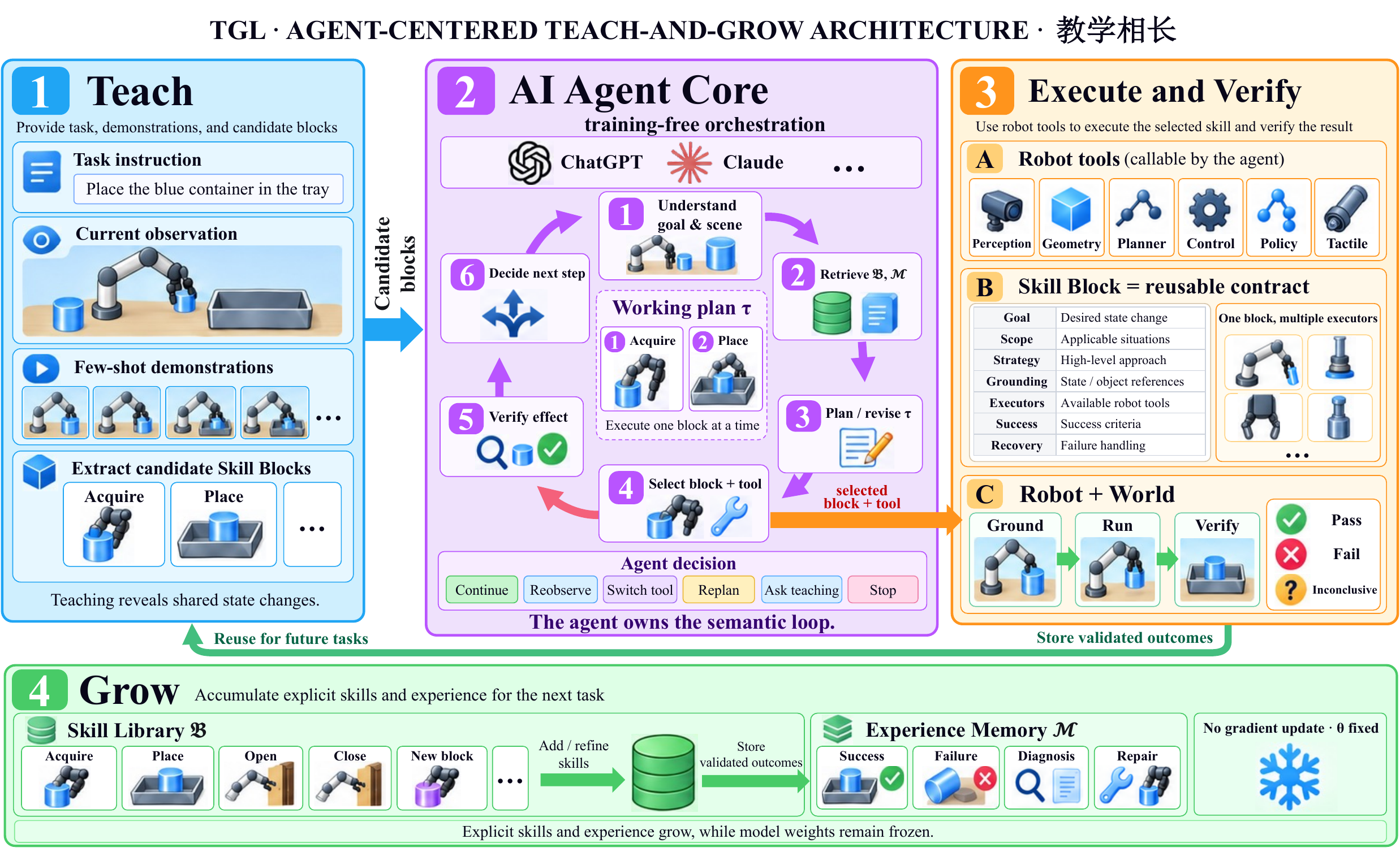}
    \caption{Teach-and-Grow for one unfamiliar task. Teaching reveals shared state changes and new Skill Blocks. The AI Agent composes a route, runs one block, and checks its effect after the robot acts. Verified capabilities enter the Skill Library; outcomes, failures, and repairs enter Experience Memory. The complete Grow state returns to composition on the next task.}
    \label{fig:pipeline}

\end{figure*}

These properties point to a different design target. Rather than making the policy responsible for every new behavior, can the behavior itself become an explicit object that the robot can name, test, and keep? Such an object separates what a demonstration teaches from what one scene imposes. The semantic part, namely what should change, in which order, under which conditions, and what counts as success, is shared by every successful way of performing a task, whereas the physical part, namely the pose, grasp, and path that were used, belongs to that scene and can be computed again from the robot's current measurements. Reuse therefore does not require reproducing the motion; it requires achieving and verifying the same effect.

That separation becomes workable once an AI agent owns the organizing role, because agents and robots already share one workflow. Coding agents such as Codex and Claude Code inspect a changing workspace, form a plan, call tools, read the result, and revise their strategy~\cite{openai2025codex,anthropic2025claudecode}. Robot autonomy has long had the same shape: perceive, plan, decide, act, and feed the consequence back into the next decision, which is how modular robot pipelines have always been organized. The difference is that an agent, rather than a hand-written controller, now owns that loop, so TGL migrates the agent system onto the robot: camera observations are the state, tools for metric geometry, collision checking, contact handling, and continuous control are the operations, and measurements of what changed are the results.

In our own experiments, an agent driving the robot through those tools completed manipulation tasks with no teaching at all, which is why TGL treats demonstrations as an accelerator rather than a precondition; that behavior, however, was not reliable enough to retain, and unguided exploration on hardware carries risk. Other systems report a related picture: zero-shot agent exploration can assemble and repair robot behavior without updating a task-specific policy~\cite{lu2026aspire,zhang2026playful}, and lifelong, memory-based systems study how skills and experience accumulate across tasks~\cite{tziafas2024lrll,wang2026skillmemo}. Because trial and error on a physical robot still requires many sequential decisions and can trigger unsafe actions, TGL seeds task structure with sparse successful teaching and reserves exploration for the gaps that remain.

Removing the retraining tax is the design goal of Teach-and-Grow Learning (TGL), an architecture for training-free skill acquisition from sparse teaching. The agent aligns several demonstrations by what they accomplish and synthesizes a shared strategy for each subgoal, recording it as a closed-loop \emph{Skill Block}, which is not a trajectory fragment to replay but a behavior whose objects, geometry, and control targets are re-instantiated from the current scene. At run time the agent grounds each block in the current scene, executes one meaningful stage, and checks the measured effect before deciding what comes next: continue, look again, repair the remaining route, or ask for one targeted demonstration. Training-free has a precise meaning here. Acquiring the incoming task involves no gradient update, no fine-tuning, and no reinforcement-learning stage. The pretrained models remain fixed, and what changes is the explicit skill and memory state. Our implementation uses OpenAI GPT-6 Astra~\cite{openai2026gpt6astra} for multimodal reasoning and Codex for interaction with robot tools.

Execution feeds two persistent stores. Blocks that pass validation enter a \emph{Skill Library}, while their conditions, outcomes, diagnoses, and repairs enter a structured \emph{Experience Memory}. Together they close a loop in which teaching becomes cheaper as the robot grows: teaching supplies the starting structure, execution exposes what the teaching left unresolved, and those gaps make the next intervention specific. This is the classical Chinese principle \raisebox{-0.18ex}{\includegraphics[height=1.35ex]{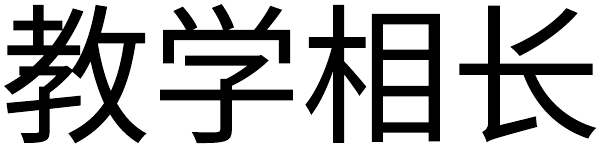}} (\emph{jiào xué xiāng zhǎng}), teaching and learning promote one another. The loop is also asymmetric: a successful demonstration teaches a \emph{strategy}, which generalizes across scenes, whereas a failure teaches a \emph{condition}, which keeps it from recurring.

The Agent decides which subgoal to pursue, what evidence is needed, and whether the last action achieved its intended effect, while robot-native executors handle metric motion, contact, and continuous control. Physical evidence connects the two levels, and because the route is explicit, components improve one at a time: a grasp detector can be replaced inside a block, and a validated block enters the library as a separately retrievable behavior.

Fig.~\ref{fig:teaser} contrasts the two routes. The end-to-end route asks the weights to absorb each new piece of the physical world, whereas TGL converts sparse evidence into reusable knowledge. It is not an adapter around a fixed policy: learned policies, planners, perception models, and controllers become capabilities that the Agent organizes.

That separation also isolates deliberation from execution. Agentic acquisition observes, reasons, and calls tools repeatedly, which an unfamiliar task needs but at a cost above feed-forward inference. TGL therefore lets a VLA, WAM, or other fast policy execute a mature block directly, and lets verified trajectories train such a student later, so a new task becomes executable as soon as its taught skills pass validation, while a familiar task runs at policy speed.

The lifecycle also changes which quantity should scale. Neural scaling laws relate loss to resources such as parameters, data, and compute~\cite{kaplan2020scaling,hoffmann2022training}. Let $X>0$ denote the robot's \emph{effective reusable experience}: accumulated interaction that has survived validation and can still be retrieved, grounded, and composed in a new scene. We propose the Teach-and-Grow scaling-law hypothesis
\begin{equation}
\begin{aligned}
    \mathcal{E}_{\mathrm{future}}(X) &= \mathcal{E}_{\infty}+A X^{-\alpha},\\
    D_{\mathrm{teach}}(X) &= D_{\infty}+B X^{-\beta},
\end{aligned}
\label{eq:intro_scaling}
\end{equation}
where $\mathcal{E}_{\mathrm{future}}$ is expected error on related future tasks, $D_{\mathrm{teach}}$ is the teaching needed to acquire one, $\mathcal{E}_{\infty},D_{\infty}\geq0$, and $A,B,\alpha,\beta>0$. Unlike model size or a frozen offline dataset, $X$ grows out of the task interactions a deployed robot already performs. The hypothesis predicts that retaining more reusable experience lowers both the error on related future tasks and the teaching required to acquire them, so deployment becomes the beginning of accumulation.

\newpage
Our contributions are:
\begin{itemize}
    \item We formulate \textbf{Teach-and-Grow Learning}, a training-free architecture that acquires manipulation skills from sparse teaching and reaches 99.9\% mean success on LIBERO and 92.4\% on LIBERO-Plus with fixed pretrained weights.
    \item We introduce reusable closed-loop \textbf{Skill Blocks} that turn several demonstrations into shared subgoal strategies while recomputing physical realization from the current scene.
    \item We keep validated behaviors in a persistent \textbf{Skill Library} and their context in structured \textbf{Experience Memory}, so a failure and its repair affect later tasks rather than only the episode that produced them.
    \item We propose a \textbf{Teach-and-Grow scaling-law hypothesis} for lifelong robot intelligence, relating effective reusable experience to future-task error and the marginal amount of teaching required for new tasks.
\end{itemize}

\section{Related Work}
\label{sec:related}

\subsection{Generalist VLA and World-Action Policies}

RT-1 and RT-2 showed that large-scale behavior cloning and web-to-robot transfer can extend robot behavior~\cite{brohan2022rt1,zitkovich2023rt2}, later generalist policies covered additional model families~\cite{octoteam2024octo,kim2024openvla,black2024pi0}, and larger robot datasets broadened their empirical base~\cite{openx2023,khazatsky2024droid,wu2024robomind}. Across these efforts, scaling data and model resources produces systematic gains~\cite{sartor2024neuralscaling,lin2024datascaling,zha2025factoredscaling}.

These results establish end-to-end scaling as a practical route, yet they also expose its binding resource: physical coverage. Probing studies report a gap between instruction understanding and reliable execution under distribution shift~\cite{fang2025intact}, and the field has responded with online reinforcement learning, few-shot policy modules, and world-action models that learn visual dynamics and actions jointly~\cite{lu2025vlarl,ding2026skillplug,ye2026dreamzero}. Those approaches reduce the adaptation burden, but they do not change where the new competence is stored.

Rather than replacing these policies, TGL organizes them within a broader learning system. Here a generalist policy can execute a block, while the Agent acquires and validates new task structure through teaching and interaction. Furthermore, the resulting explicit behaviors can later supply trajectories for fast-policy distillation.

\subsection{Demonstrations, Few-Shot Adaptation, and Reusable Skills}

Learning from demonstration supplies successful structure without unrestricted physical exploration: existing systems use demonstrations, play, and language to generate behavior that can be transferred or recombined~\cite{mandlekar2023mimicgen,xu2023xskill,zhu2025atomskill}, and a closely related line makes skill structure explicit for continual learning, symbolic composition, or few-shot adaptation~\cite{zhang2026sce,quartey2026pacts,ding2026skillplug}.

In TGL, by contrast, demonstrations support the inference of semantic subgoals, reusable relations, ordering, expected effects, and applicability, and a Skill Block can contain several perception-action iterations using a learned policy, planner, servo, controller, or tool composition. Its strategy is grounded in current observations at execution time, and its scope, executor contracts, and outcome test are explicit objects that teaching can create and validation can inspect.

\subsection{Agentic Tool Use, Memory, and Lifelong Robots}

Tool-using agents have shown that high-level reasoning can organize robot programs, constraints, and feedback~\cite{ahn2022saycan,liang2022codeaspolicies,huang2024rekep}. Recent robot operating layers extend this idea to skill graphs, memory, and cross-embodiment interfaces~\cite{liu2026phyagentos,qin2026aeros,yoon2026robobridge}. Other systems retain deployment experience or grow and repair skill libraries over time~\cite{wu2025dejavu,tziafas2024lrll,lu2026aspire}.

Episodic memory, tools, and executable skills play distinct roles: memory records what happened and why an attempt failed, a tool provides an operation the robot can invoke, and an executable skill connects an intended change to current-scene grounding, an executor, and an outcome test. Storing more episodes therefore increases context, while adding a validated skill expands what can be composed, which is why TGL uses memory to guide selection and Skill Blocks to represent what the robot can execute, instead of promoting every episode to a new behavior.

TGL brings these threads together through executable skills acquired without task-specific training: demonstrations give the agent a subgoal strategy it can test and retain, and the resulting block and its experience then guide later tasks, so each teaching interaction can contribute behavior beyond its original episode.

\section{Teach-and-Grow Learning}
\label{sec:formulation}

\subsection{Learning with Fixed Model Weights}

Training-free task acquisition in TGL expands explicit skills and memory while keeping the pretrained agent and robot models fixed. Demonstrations serve as reasoning inputs, and no optimizer is invoked for the new task. Consider a sequence of tasks $\{\mathcal{T}_n\}_{n=1}^{N}$. When task $\mathcal{T}_n$ arrives, the robot receives a small teaching set
\begin{equation}
    \mathcal{D}_n=\{d_n^{(1)},\ldots,d_n^{(m_n)}\}, \qquad m_n\ll |\mathcal{D}_{\mathrm{pretrain}}|,
\end{equation}
where $\mathcal{D}_{\mathrm{pretrain}}$ denotes the corpus used to train the fixed foundation stack. The robot begins with a Skill Library $\mathcal{B}_n$, Experience Memory $\mathcal{M}_n$, and interaction history $\mathcal{H}_n$. Let $\Delta\mathcal{H}_n$ contain the useful lessons from executing this task. The update is
\begin{equation}
\begin{aligned}
    \theta_{n+1}&=\theta_n=\theta,\\
    (\mathcal{B}_{n+1},\mathcal{M}_{n+1})
    &=\mathcal{U}(\mathcal{B}_n,\mathcal{M}_n,\mathcal{D}_n,\Delta\mathcal{H}_n),\\
    \mathcal{H}_{n+1}&=\mathcal{H}_n\cup\Delta\mathcal{H}_n .
\end{aligned}
\label{eq:tgl_update}
\end{equation}
New behaviors enter $\mathcal{B}$, and their context, failures, and repairs enter $\mathcal{M}$. With $\theta$ fixed, learning changes the behaviors the robot can retrieve and execute.

Accordingly, Eq.~\ref{eq:tgl_update} assigns a separate role to each part of the persistent state. The fixed parameters $\theta$ provide pretrained semantic and physical priors. The Skill Library $\mathcal{B}$ changes when a block passes its admission checks. Experience Memory $\mathcal{M}$ records task context, outcomes, diagnoses, and repairs to guide later decisions. The history $\mathcal{H}$ preserves the underlying evidence, and $X$ summarizes the part that remains verified and reusable. A future fast student has separate parameters $\phi$ and can be trained on verified trajectories. Thus, task acquisition updates explicit skills and memory with $\theta$ fixed; later distillation updates $\phi$ to compress the acquired behavior.

By contrast, behavior cloning turns demonstrations into supervised action targets, and reinforcement learning uses interaction to change a policy or value function. TGL treats demonstrations as evidence about task structure and makes executable capability itself the object of learning.

\subsection{Skill Blocks}

A Skill Block is represented as
\begin{equation}
    b_i\mathrel{:=}\langle g_i,\mathcal{S}_i,\rho_i,\gamma_i,\Pi_i,v_i,\mathcal{R}_i\rangle .
\label{eq:skillblock}
\end{equation}
Each field answers one practical question: $g_i$ says what should change; $\mathcal{S}_i$ says when the block applies; $\rho_i$ stores the reusable strategy; $\gamma_i:\mathcal{O}\rightarrow\mathcal{Z}_i$ maps the observation space $\mathcal{O}$ to the block's grounded variables $\mathcal{Z}_i$; $\Pi_i$ lists compatible executors; $v_i$ maps pre-action observations, post-action observations, and executor evidence to \emph{pass}, \emph{fail}, or \emph{inconclusive}; and $\mathcal{R}_i$ lists bounded recovery choices. At decision step $t$, the Agent selects an executor $\pi_{i,t}\in\Pi_i$.

The word \emph{block} emphasizes composition rather than a low-level motor primitive: a block for acquiring the requested container turns a semantic request into a grounded grasp and closes its loop by verifying retention, and the next block needs only the achieved effect, not the backend that produced it.

A complete task is represented at each semantic decision step by an ordered working plan
\begin{equation}
    \tau_t=\big(b_{i_{t,1}},b_{i_{t,2}},\ldots,b_{i_{t,L_t}}\big),
\label{eq:block_comp}
\end{equation}
where $L_t$ is the current route length. After a block returns an outcome, the agent may keep, shorten, or replace the remaining route. The tuple represents the remaining plan at the current decision step.

\subsection{Structured Experience Memory}

A Skill Block stores a behavior; Experience Memory stores the lesson learned from using it. Each record is a structured textual item
\begin{equation}
\begin{aligned}
    \mu_j=\langle &\text{task},\text{context},\text{blocks},\text{outcome},\\
                 &\text{diagnosis},\text{repair},\text{evidence}\rangle .
\end{aligned}
\label{eq:experience_memory}
\end{equation}
Each record describes the outcome, its diagnosis, and any successful repair, retains links to observations and provenance, and is retrieved as compact guidance that a person can inspect or revise.

\subsection{The Teach-and-Grow Scaling-Law Hypothesis}

The scaling resource in TGL measures how much past experience remains useful for action in a new scene. At a checkpoint with at least one positively weighted experience, define
\begin{equation}
    X_n=\sum_{h\in\mathcal{H}_n}\omega(h;\mathcal{B}_n,\mathcal{M}_n),
\label{eq:effective_experience}
\end{equation}
where $\omega(h;\mathcal{B}_n,\mathcal{M}_n)\in[0,1]$ is a dimensionless score fixed before evaluation; the positive-weight condition above ensures $X_n>0$. The score uses only evidence available at checkpoint $n$: evidence reliability, added coverage, retrievability, grounding validity, and compatibility with admitted blocks. Future-task outcomes never enter it. Thus, two robots can store the same number of episodes yet possess very different $X_n$. In Eq.~\ref{eq:intro_scaling}, $X$ denotes this same checkpoint quantity under the fixed scoring rule.

Equation~\ref{eq:intro_scaling} states the central prediction. As effective experience grows, error on related future tasks and the teaching required to acquire them should fall toward task-dependent floors. The exponents $\alpha$ and $\beta$ characterize how efficiently a system turns lived experience into future competence. A robot that records many episodes but cannot retrieve or re-ground them will scale poorly; one that turns a small number of verified episodes into broadly reusable blocks may scale well.

The law can be tested over sequential acquisition experiments: hold the foundation models and tools fixed, grow $X$, and measure future-task error and marginal teaching at each checkpoint. We define $D_{\mathrm{teach}}$ as teacher intervention time under a fixed protocol until the preregistered success criterion is reached. If the criterion is not reached before the preregistered teaching budget is exhausted, the observation is recorded at that cap rather than omitted. Because Skill Blocks remain explicit objects, the same study can also track whether new growth preserves earlier behavior.

\section{General Agentic Robot Architecture}
\label{sec:method}

Fig.~\ref{fig:pipeline} follows one unfamiliar task from teaching to reuse: demonstrations provide the subgoal structure, the agent selects blocks and robot tools, checks the result of each action, and retains useful behavior and experience for later tasks. The architecture defines how the agent, reusable blocks, and executors exchange decisions and physical evidence, and the choice of reasoning model, agent runtime, and robot tools is left to the implementation.

Within that architecture, the Agent interprets the task, retrieves candidate blocks, and decides what remains to be done. Each selected block identifies the relevant objects and geometry in current observations before its executor acts. Fresh observations and executor feedback then determine whether the intended effect occurred. Before a block enters the library, validation checks its supported scope, executor compatibility, outcome test, recovery behavior, and provenance. The same contract supports later improvements: a detector, planner, controller, or learned policy can be replaced and tested within the block.

\subsection{Few-Shot Teaching and Cross-Demonstration Abstraction}

For a new task, a few successful demonstrations show the state changes needed to reach the goal. The learner identifies events such as acquiring an object, opening a drawer, or placing one object inside another. These events define candidate subgoals for the new task.

The AI agent reads the instruction together with synchronized observations. Each demonstration is then viewed as an ordered sequence of segments, divided around meaningful changes in task state:
\begin{equation}
    d^{(j)}=\left(s_1^{(j)},s_2^{(j)},\ldots,s_{K_j}^{(j)}\right).
\end{equation}
Segments are aligned by what they accomplish, even when their timing and motion differ. Let $\mathcal{A}_k$ collect all segments aligned to semantic effect $k$. The agent then synthesizes the shared strategy
\begin{equation}
    \rho_k=\operatorname{Synthesize}(\mathcal{A}_k).
\label{eq:synthesis}
\end{equation}
The agent keeps the shared roles, relations, ordering, and effects, then recomputes scene-specific details such as pose, grasp, path, and control. Differences among demonstrations show how widely the block can be reused. A pattern seen with one instrument stays narrow; a strategy repeated across interchangeable objects can support a broader scope.

\subsection{Skill Blocks in the Current Scene}

\begin{figure*}[!t]
    \centering
    \includegraphics[width=0.90\textwidth]{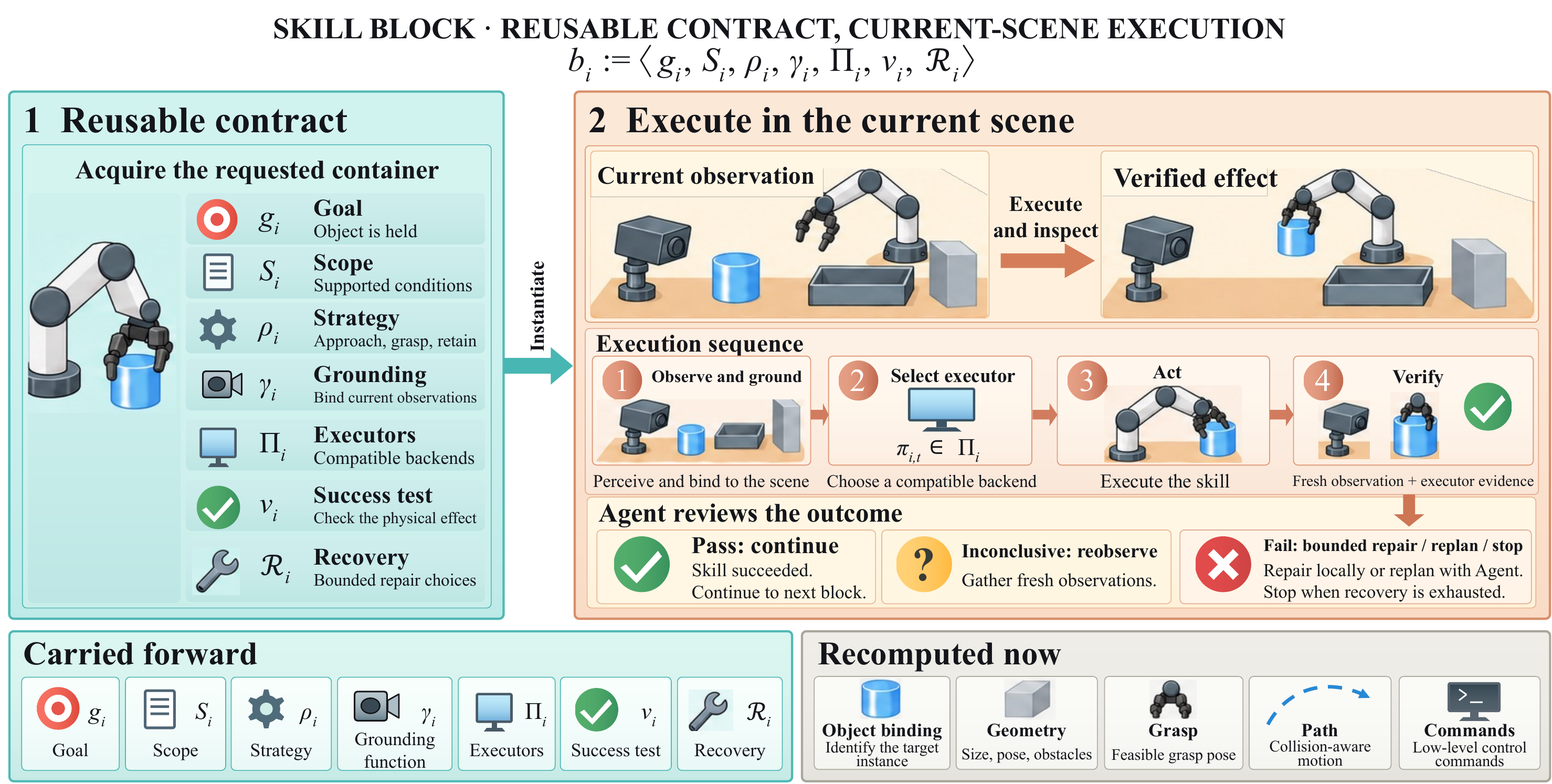}
    \caption{A Skill Block is a reusable contract for one meaningful change. It carries a goal, scope, strategy, grounding function, executor set, success test, and recovery. At runtime it observes the scene, binds current objects and geometry, chooses an executor, acts, and verifies the result.}
    \label{fig:skillblock}
    
\end{figure*}

Fig.~\ref{fig:skillblock} shows how a block combines a reusable strategy with current observations: its goal and strategy persist across tasks, while object poses, grasp candidates, and motions are computed for the scene in front of the robot.

A new block is then evaluated in scenes separate from its teaching demonstrations. Successful executions define the supported range, while failures guide refinement or a narrower scope. Only after validation does the block enter the library, and its scope travels with it, so a behavior is never offered for a scene it was not shown to handle.

Crucially, different backends may realize the same block as long as they achieve and verify the same result. An acquisition block may use geometric grasping, a VLA, or a tactile policy. A block is thus closer to a contract than to a script: it states what must hold before and after, and leaves the means of achieving it open. Adding a new sensor or executor therefore changes only the blocks that need it.

\subsection{Dynamic Agentic Composition and Tool Selection}

Given a task instruction, current observations, and persistent state $(\mathcal{B}_n,\mathcal{M}_n)$, the Agent retrieves a compact set of blocks and forms the working plan in Eq.~\ref{eq:block_comp}. It selects executors according to the current evidence. A familiar scene may use a learned policy, while uncertain contact may call for geometry, touch, or a more deliberate controller. The Agent can revise this choice at each semantic state transition.

Continuous motor control runs within the selected executor. At semantic decision step $t$, a block returns the next observation and outcome evidence:
\begin{equation}
    (o_{t+1},\varepsilon_t)
    =\operatorname{Run}(b_{i_{t,1}},\pi_{i_{t,1},t},o_t).
\end{equation}
Here $\varepsilon_t$ records executor and verifier evidence, including whether the intended effect passed, failed, or remained inconclusive. The agent uses that evidence to continue, look again, choose another route, ask for teaching, or stop, so the remaining plan follows what the robot actually did.

Task-level decisions can change both the remaining subgoals and the tools used to reach them. The Agent may request another sensor observation, call a geometric planner, retrieve a different behavior, or replace part of the block sequence. The executor continues to handle its local action loop and returns evidence at semantic boundaries.

\subsection{Physical Feedback, Bounded Exploration, and Targeted Teaching}

After a block executes, the agent compares the observed effect with the intended change. If they agree, the task continues. If they differ, the agent looks again, selects another executor or route, or asks the teacher for a focused demonstration. Once a mismatch is detected, the plan is corrected before the next block runs, so one failed effect does not propagate through the remaining route.

Targeted teaching broadens this loop. The agent can ask for another demonstration when one would introduce a missing behavior, clarify an ambiguous subgoal, or move acquisition forward faster than unguided trial. Requests of this kind concentrate the teacher's effort on the gap that execution exposed, and they become less frequent as the library covers more of the task. Teaching is therefore an accelerator rather than a precondition: the same loop can run with no demonstrations at all, at the cost of more trials and more model calls.

\subsection{Structured Growth and Fast-Policy Distillation}

Validated reusable behavior enters the Skill Library; Experience Memory records outcomes, diagnoses, and repairs. Both stores are searchable, versioned, and human-editable, so later tasks can reuse a behavior together with the context that supports its selection.

Verified trajectories can also train a fast student: a VLA or WAM may execute a mature block directly, while the agentic path continues to handle novelty, diagnosis, and capability expansion. As robot foundation models become stronger, several specialist tools may collapse into a single executor, and the Agent then coordinates fewer components without changing the interface that teaching and validation use.

\section{Experiments and Results}
\label{sec:experiments}

We evaluate TGL on LIBERO~\cite{liu2023libero} and LIBERO-Plus~\cite{fei2025liberoplus}. LIBERO-Plus extends the original benchmark with perturbations to object layouts, camera viewpoints, robot initial states, instructions, lighting, backgrounds, and sensor observations. The benchmark evaluations assess task success, while controlled studies examine how demonstrations become executable blocks, how those blocks persist, and how feedback changes the plan. Appendix~\ref{app:experiments} gives the controlled-study protocols and trace provenance.

Tables~\ref{tab:benchmark_results} and~\ref{tab:plus_results} summarize the resulting success rates.

On LIBERO, TGL reaches 99.9\% mean success, matching LaST-R1 for the highest mean among the compared methods. It exceeds OpenVLA-OFT by 2.8 percentage points and SRPO by 0.7 points, while acquiring its skills without policy optimization. Object and Goal reach 100.0\%, and Long reaches 99.9\%.

On LIBERO-Plus, TGL obtains the highest category mean, 92.4\%, compared with 89.7\% for $\pi_{0.5}$ with RAS and MCSI, 85.8\% for InternVLA-A1.5, and 84.5\% for VLANeXt. TGL leads on robot initial states, language, lighting, and layout. Its camera and sensor-noise scores identify the perturbations for which stronger visual grounding would be most useful. The controlled studies below examine how taught skills and physical feedback contribute to execution.

\begin{table*}[!t]
\caption{LIBERO success rates (\%). Mean averages the four suites equally. Bold marks the best result in each column, including ties.}
\label{tab:benchmark_results}
\centering
\begin{tabular}{@{}lrrrrr@{}}
\toprule
Method & Spatial & Object & Goal & Long & Mean \\
\midrule
OpenVLA~\cite{kim2025openvlaoft} & 84.7 & 88.4 & 79.2 & 53.7 & 76.5 \\
OpenVLA-OFT~\cite{kim2025openvlaoft} & 97.6 & 98.4 & 97.9 & 94.5 & 97.1 \\
SRPO~\cite{fei2025srpo} & 98.8 & \textbf{100.0} & 99.4 & 98.6 & 99.2 \\
VLANeXt~\cite{wu2026vlanext} & 99.0 & 99.2 & 96.6 & 94.8 & 97.4 \\
InternVLA-A1.5~\cite{internrobotics2026a15} & 98.6 & 99.8 & 98.6 & 98.4 & 98.9 \\
LaST-R1~\cite{chen2026lastr1} & \textbf{99.8} & \textbf{100.0} & \textbf{100.0} & 99.8 & \textbf{99.9} \\
\textbf{TGL (ours)} & 99.7 & \textbf{100.0} & \textbf{100.0} & \textbf{99.9} & \textbf{99.9} \\
\bottomrule
\end{tabular}
\end{table*}

\begin{table*}[!t]
\caption{LIBERO-Plus success rates (\%). Mean averages the seven perturbation categories equally. Bold marks the best result in each column.}
\label{tab:plus_results}
\centering
\begin{tabular}{@{}lrrrrrrrr@{}}
\toprule
Method & Camera & Robot & Language & Light & Background & Noise & Layout & Mean \\
\midrule
OpenVLA~\cite{fei2025liberoplus} & 0.8 & 3.5 & 23.0 & 8.1 & 34.8 & 15.2 & 28.5 & 16.3 \\
$\pi_0$~\cite{fei2025liberoplus} & 13.8 & 6.0 & 58.8 & 85.0 & 81.4 & 79.0 & 68.9 & 56.1 \\
OpenVLA-OFT~\cite{fei2025liberoplus} & 56.4 & 31.9 & 79.5 & 88.7 & 93.3 & 75.8 & 74.2 & 71.4 \\
OpenVLA-OFT+~\cite{fei2025liberoplus} & \textbf{92.8} & 30.3 & 85.8 & 94.9 & 93.9 & 89.3 & 77.6 & 80.7 \\
VLANeXt~\cite{wu2026vlanext} & 90.4 & 65.7 & 81.8 & 95.9 & 82.5 & 94.1 & 80.8 & 84.5 \\
InternVLA-A1.5~\cite{internrobotics2026a15} & 83.1 & 55.1 & 86.9 & 96.4 & \textbf{98.2} & \textbf{95.6} & 85.2 & 85.8 \\
$\pi_{0.5}$ + MCSI~\cite{zhan2026stable} & 86.0 & 83.0 & 83.0 & 97.0 & 98.0 & 92.0 & 87.0 & 89.4 \\
$\pi_{0.5}$ + RAS + MCSI~\cite{zhan2026stable} & 81.0 & 88.0 & 90.0 & 97.0 & 96.0 & 90.0 & 86.0 & 89.7 \\
\textbf{TGL (ours)} & 87.3 & \textbf{90.7} & \textbf{96.8} & \textbf{97.3} & 97.4 & 89.9 & \textbf{87.2} & \textbf{92.4} \\
\bottomrule
\end{tabular}
\end{table*}

\begin{table*}[!t]
\caption{Controlled Teach-and-Grow studies on LIBERO. Detailed splits and statistics appear in Appendix~\ref{app:experiments}.}
\label{tab:results}

\centering
\renewcommand{\arraystretch}{1.15}
\setlength{\tabcolsep}{6pt}
\begin{tabular}{@{}p{0.23\textwidth}p{0.32\textwidth}p{0.36\textwidth}@{}}
\toprule
Study & Observed result & Main lesson \\
\midrule
Task-specific learning cycle & Two learned blocks solve 3/3 states, survive save-and-reload at 3/3, and stop at the first unmet semantic effect outside their learned scope & Demonstrations become persistent, scope-aware behavior \\
Feedback-driven execution & Two representative successful traces replan or reobserve after physical feedback & Outcomes change what the Agent does next \\
Fixed-executor library pilot & Six-block library: 0/6; eight-block library: 4/6 & Local library growth changes what the same executor do \\
\bottomrule
\end{tabular}

\end{table*}

\paragraph{From demonstrations to persistent blocks.}
Ten visual demonstrations decompose into acquisition and release stages, and all 20 observable effects are confirmed. In a separate learning cycle, three teacher trajectories produce two blocks: acquire the requested object and release it in the required relation. The pair solves all three evaluation states, survives save-and-reload, and solves the same states again. Save-and-reload shows that the taught route has joined the library as a reusable alternative; the existing six-block route remains available on the same states. Beyond its learned range, the route stops when the expected effect is absent, so one mistake does not propagate through the task.
The reported visual study uses a deterministic decomposition; the general formulation can place a multimodal Agent in this role.

\paragraph{Feedback changes the workflow.}
For the bowl-on-plate task, the Agent begins with an acquisition block followed by a release block. The pick succeeds, but the post-grasp evidence is not yet sufficient for placement. The Agent rebuilds the remaining route and completes the task. For the drawer-opening task, an inconclusive outcome prompts a fresh observation. In both traces, evidence from the physical world changes the Agent's next action.

\paragraph{Local growth under a fixed executor.}
The six-block library solves 0/6 evaluations; adding the two learned blocks raises success to 4/6. Model weights, runtime, and evaluation budget remain fixed. This two-task pilot illustrates how library content can change the behavior available to the same executor. The corresponding uncertainty estimates and acquisition costs are reported in Appendix~\ref{app:experiments}.

\begin{figure*}[!t]
    \centering
    \includegraphics[width=0.9\textwidth]{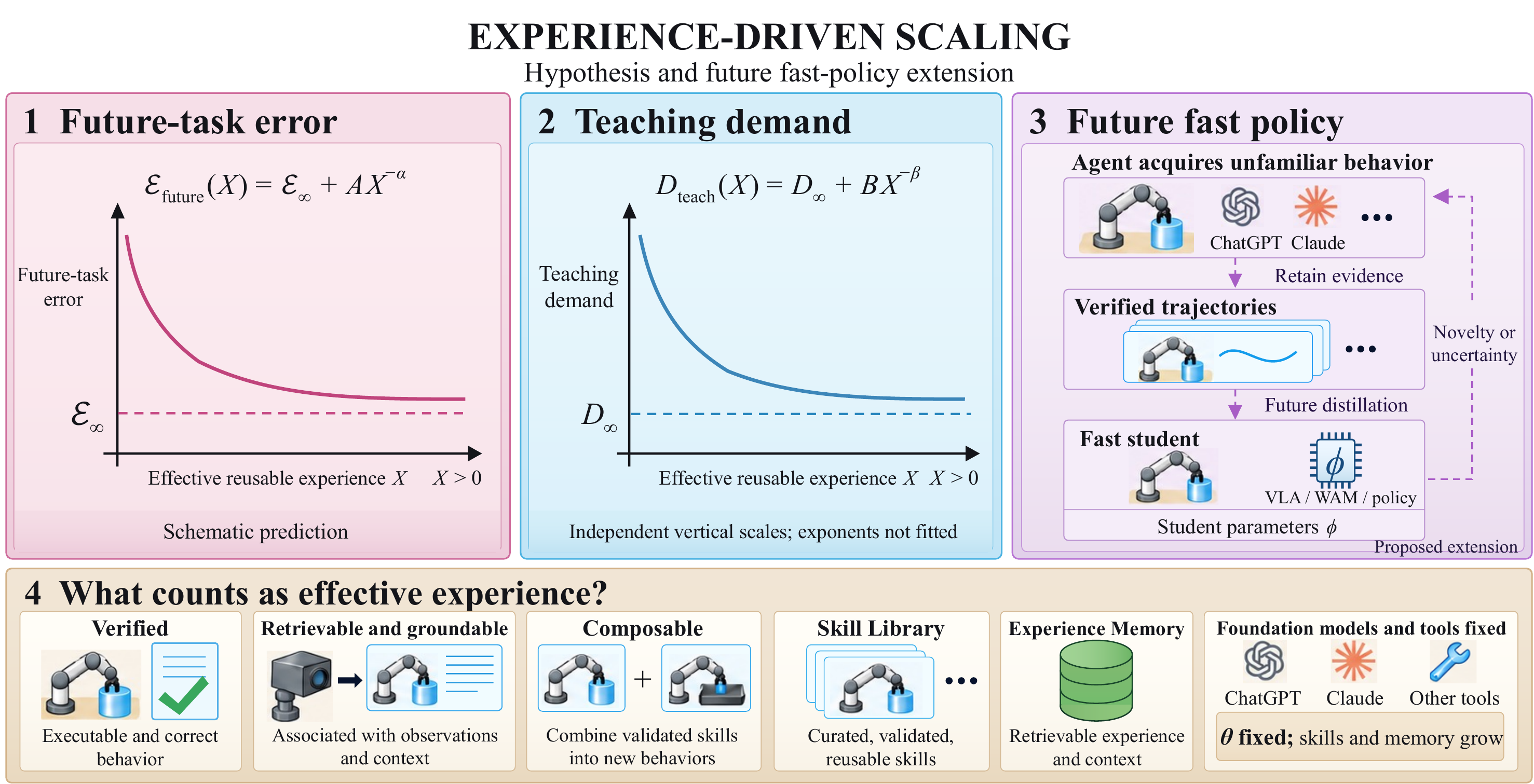}
    \caption{Experience-driven robot scaling. The two aligned plots use separate schematic vertical scales: as effective reusable experience $X$ grows, future-task error and teaching demand are predicted to approach their own irreducible floors. The Agent learns unfamiliar tasks and keeps verified experience; a future compression path moves mature behavior into a fast policy, with uncertain cases returning to the Agent.}
    \label{fig:scaling}
    
\end{figure*}

\section{Discussion}
\label{sec:discussion}

\subsection{Learning as Capability Growth}

Most learning systems store new ability inside a policy, value function, or world model. TGL stores it a second way as well: as reusable behavior, together with the experience that explains it. Because that behavior is explicit, it counts as learned knowledge only once the robot can recognize where it applies and produce the intended result in a new scene.

Few-shot teaching supplies the subgoal order and success conditions, and the robot computes geometry and control from the current scene, so exploration narrows while execution still reveals repairs beyond the demonstrated trajectories.

\subsection{Persistent Memory and Non-Destructive Growth}

A new task leaves behind a named behavior and the context that makes it useful: when a related task arrives, the Agent retrieves that behavior, checks its applicability, and grounds it in the new scene.

Experience records explain why a block was selected, which effect was expected, and what the robot actually observed, and they connect a failure to the decision that produced it. The executable response and the diagnostic record stay separate: a revised behavior enters the library only after validation, while its supported conditions and outcome remain in memory. Correction therefore stays local: a failed effect identifies the behavior or observation that needs attention, earlier skills remain individually addressable, and growth is additive rather than destructive, because the robot can extend its repertoire without reopening the parts that already work.

\subsection{The Agent as Organizing Intelligence}

At each semantic boundary, the Agent checks whether the remaining plan still fits the observed state, requesting new evidence or replacing the route while executors handle motion; the two LIBERO traces show both decisions.

Perception models, geometric planners, servos, controllers, and VLA policies provide the physical capabilities this reasoning acts on. A stronger executor may combine several of them; the Agent then works with fewer tools while still selecting subgoals, inspecting outcomes, and retaining useful behavior.

\subsection{The Teach-and-Grow Scaling-Law Hypothesis}

Model scaling improves the prior before deployment, and test-time scaling spends more computation on one episode. Fig.~\ref{fig:scaling} adds a third, lifetime scaling: useful experience from one task changes how the next task begins. A single bridge block can open many new compositions, whereas a large archive of redundant trajectories may add almost nothing, so what matters is not how much the robot has stored but how much of it remains verified and reusable.

With foundation models and tools fixed, Eqs.~\ref{eq:intro_scaling} and~\ref{eq:effective_experience} predict declining future-task error and teaching demand as $X$ grows. Appendix~\ref{app:scaling} defines checkpoints and cost accounting for testing these predictions.

\subsection{A Slow Teacher and a Fast Student}

In the proposed two-speed extension, the agentic route learns unfamiliar tasks while a distilled fast policy executes familiar ones; verified trajectories can train a VLA, WAM, diffusion policy, or block-specific controller, and when the fast path is uncertain, control returns to the Agent, which diagnoses the gap and expands the library.

The same division can later extend to a fleet: robots may share candidate strategies and failure signatures, while each receiver re-grounds and verifies them locally.

\subsection{Execution Cost}

Sequential reasoning and repeated observations add latency to unfamiliar tasks, so TGL places those decisions at subgoal boundaries, with executors handling continuous control below them. Whether a stored skill stays useful as the library grows then depends on retrieval quality and grounding.

\section{Conclusion}
\label{sec:conclusion}

Teach-and-Grow Learning turns a few successful demonstrations into reusable robot capability while the pretrained models stay fixed. OpenAI GPT-6 Astra reasons over the task and physical feedback, and Codex connects its decisions to robot tools. The implementation reaches 99.9\% mean success on LIBERO and 92.4\% on LIBERO-Plus, matching the highest LIBERO mean and the highest perturbation-category mean among the compared methods.

The controlled studies connect those numbers to concrete learning operations: demonstrations produce executable blocks that survive save-and-reload, and two additional blocks raise related-task success with the same executor and budget. Reusable experience then lowers future-task error and teaching demand, and mature behavior can supply trajectories for a fast student. Deployment therefore has a continuing learning role: the robot acquires a skill through teaching, puts it to work with fixed weights, and adapts it to later tasks.

\label{page:main-end}
\clearpage
\bibliographystyle{IEEEtran}
\bibliography{references}

\FloatBarrier
\appendices

\section{Why End-to-End Scaling Remains Costly}
\label{app:endtoend}

\subsection{Collecting Physical Interaction Data}

Large language models began with an unusual historical advantage: vast quantities of text, code, books, and webpages had already been produced for human purposes. Model builders still pay heavily to curate and train on that material, but they do not need to recreate every sentence by operating a physical machine. General robot learning faces a different starting condition. Much of its valuable evidence consists of synchronized observations, actions, embodiment state, contact, calibration, failure, recovery, and final physical effect. This evidence does not exist until a robot or simulator is run. Open X-Embodiment, DROID, RoboMIND, and AgiBot World are major advances~\cite{openx2023,khazatsky2024droid,wu2024robomind,agibot2025world}. Their scale also shows how much new infrastructure and interaction must be created before robot data can resemble a mature web-scale substrate.

The most consequential evidence is often the least likely to appear in a successful demonstration set: near collisions, transparent or deformable objects, unstable grasps, and the chained states produced by earlier mistakes. These events determine whether a robot is useful outside a curated distribution. A broad visual backbone can contribute language, object semantics, and strong priors, yet it does not automatically supply embodiment-specific geometry, contact dynamics, or action-conditioned consequences~\cite{fang2025intact}. Robot generality therefore depends on both model capacity and the cost of collecting representative physical interaction.

\subsection{Coverage Grows by Interacting Factors}

The difficulty, however, is not captured by a single count of demonstrations. Let a deployment domain contain $F$ physically relevant factors, including object identity, pose, clutter, illumination, camera, gripper, material, contact mode, and task relation. If $n_j$ denotes the meaningfully different regimes for factor $j$, a dense Cartesian design contains
\begin{equation}
    N_{\mathrm{dense}}=\prod_{j=1}^{F} n_j
\label{eq:dense_coverage}
\end{equation}
coverage cells, where $n_j\geq1$. This idealized product shows why physical coverage can grow much faster than any one dataset coordinate: adding another independent factor multiplies the joint space. Even pairwise screening must consider $F(F-1)/2$ factor pairs, with $\sum_{j<k}n_jn_k$ joint pairwise cells. Factored scaling curves were introduced precisely because exhaustive environmental variation is prohibitively expensive~\cite{zha2025factoredscaling}.

Precision creates a second pressure. Let $P$ denote a target tolerance, where a smaller value is stricter and $c$ is the smallest attainable tolerance. One recent analysis reports a regime of the form $\log N_{\mathrm{demo}}\propto1/(P-c)$ for $P>c$ as $P\downarrow c$~\cite{xu2026precision}. This relation shows why producing the final increments of physical precision can require sharply more data.

\subsection{The Retraining Tax Is Global Even When Failure Is Local}

Suppose a policy fails only for one unfamiliar container under one camera and grasp configuration. The missing competence is local, but repairing a monolithic policy commonly reopens a global process: collect representative interaction, optimize a new version, and recheck previously supported behavior. Adapters, replay, modular heads, online reinforcement learning, and continual-learning methods can reduce this burden, but they do not remove the systems fact that a change distributed through model parameters is difficult to inspect, delimit, and certify locally. As the policy's coverage expands, the cost of deciding what else may have changed expands with it.

The \emph{retraining tax} arises when a local gap requires updating and revalidating the whole policy. Teach-and-Grow changes the unit of update: a missing behavior can be added as a separately addressable Skill Block, tested where it applies, and linked to existing blocks. Appendix~\ref{app:scaling} develops the scaling and cost consequences of this change.

\subsection{The Modality and Embodiment Tax}

A second, quieter tax appears at the interfaces. Adapting a VLA to touch, audio, force, point clouds, or a new camera requires an observation interface and aligned interaction data, and a new gripper or robot likewise requires an appropriate action interface, calibration, and validation. In TGL, new sensing enters only the blocks that consume its evidence, and a new embodiment is supported through compatible executors and grounding. Integration is therefore tested at those interfaces and at the compositions they affect.

\subsection{The Long Tail Is Where Explicit Knowledge Matters Most}

End-to-end policies are strongest when deployment resembles their training data, which is precisely why the long tail deserves separate attention. Long-tail failures are sparse, varied, and often useful only after diagnosis. A failed grasp matters more when the robot records that a transparent wall confused depth, a side approach caused collision, or retention must be checked before transport. Experience Memory keeps these lessons; Skill Blocks keep the response. When retrieval, grounding, or control fails, the robot can point to what it believed, what it expected, and what local object should change.

\section{Teach-and-Grow Learning versus Existing Learning Forms}
\label{app:learningforms}

\begin{table*}[!t]
\caption{Different learning objects and update mechanisms.}
\label{tab:learning_forms}

\centering
\renewcommand{\arraystretch}{1.15}
\setlength{\tabcolsep}{5pt}
\begin{tabular}{@{}p{0.16\textwidth}p{0.22\textwidth}p{0.22\textwidth}p{0.31\textwidth}@{}}
\toprule
Property & Behavior cloning / imitation & Reinforcement learning & Teach-and-Grow Learning \\
\midrule
Primary evidence & Expert observation--action pairs & Interaction, reward, transitions & Demonstrations, instructions, success, failure, diagnosis, recovery \\
Learned object & Parametric policy & Policy, value function, or world model & Explicit Skill Library and task compositions \\
Demonstration role & Target behavior to imitate & Initialization, prior, or guidance & Seed evidence for semantic structure and reusable strategy \\
Exploration & Usually limited & Central & Targeted after few-shot teaching; bounded by safety and missing knowledge \\
Failure signal & Distribution mismatch or supervised error & Reward / return & Structured cause, negative condition, recovery, or missing block \\
Update & Gradient optimization & Gradient or search-based optimization & Create, narrow, merge, replace, compose, or retire Skill Blocks \\
Persistence & Distributed in parameters & Distributed in parameters / replay & Explicit, versioned, human-readable behavior objects \\
\bottomrule
\end{tabular}
\end{table*}

\subsection{The Learned Object and the Update Operator}

Table~\ref{tab:learning_forms} compares the learned objects and update mechanisms. In TGL, demonstrations define an initial hypothesis about semantic effects, ordering, and reusable relations. Current objects, geometry, grasps, paths, and commands remain runtime variables. The robot can realize a taught structure through different backends, acquire a recovery absent from the demonstrations, and update its explicit behavior state while the foundation policy remains fixed.

The difference can be expressed through the update operator. Parametric learning changes a model,
\begin{equation}
    \theta_{n+1}=\operatorname{Optimize}(\theta_n,\mathcal{D}_n),
\end{equation}
whereas Teach-and-Grow applies Eq.~\ref{eq:tgl_update} to the explicit skill, memory, and history state while the foundation parameters $\theta$ stay fixed. The two operations are compatible. A learned policy may implement a block, and verified behavior may later be distilled into separate student parameters $\phi$. Explicit acquisition and parametric training can therefore operate at different stages of the same learning system.

\subsection{Agent Decisions over the Evolving Task Plan}

Language and vision-language models already organize robot behavior as skill selectors, program generators, spatial reasoners, and constraint builders. SayCan ranks available skills, Code as Policies writes programs against robot APIs, VoxPoser and ReKep convert language and images into spatial objectives or constraints, and Inner Monologue incorporates environmental feedback into planning~\cite{ahn2022saycan,liang2022codeaspolicies,huang2023voxposer,huang2024rekep,huang2022innermonologue}. These systems provide the planning, tool-use, and feedback mechanisms on which later agentic architectures build.

In TGL, the agent decides how the task plan should evolve during execution. It chooses evidence, blocks, and tools; checks whether the last action achieved its effect; and revises the remaining plan. It also selects experience to retain and candidate behaviors to validate. A coding agent uses a workspace in a similar way, repeatedly inspecting state, acting through tools, and updating the plan and stored artifacts. For the robot, observations describe the physical scene, tools execute actions, and persistent artifacts contain reusable skills and experience.

Recent systems such as LRLL, ASPIRE, SkillMemo, SCE, and PACTS study lifelong skill acquisition, agentic discovery, memory, and compositional reuse~\cite{tziafas2024lrll,lu2026aspire,wang2026skillmemo,zhang2026sce,quartey2026pacts}. TGL belongs to this family. It connects few-shot abstraction, closed-loop execution, and persistent learning: teaching provides a subgoal strategy, execution tests its composition, and the resulting experience updates the Skill Library and Experience Memory. Mature behavior can later supply data for fast-policy distillation.

\subsection{Teaching and Learning Promote One Another}

The phrase \raisebox{-0.18ex}{\includegraphics[height=1.35ex]{jiaoxuexiangzhang.pdf}} comes from the classical Chinese educational text \emph{Xue Ji} and is commonly translated as \emph{teaching and learning promote one another}. In TGL, the teacher demonstrates a useful task structure. Execution then reveals what needs clarification: an object role, a recovery, an embodiment-specific requirement, or an outcome test. That evidence informs the next teaching intervention.

The resulting loop is
\begin{equation}
\text{teach}_n\rightarrow\text{compose}_n\rightarrow
\text{act{+}check}_n\rightarrow\text{grow}_n\rightarrow\text{teach}_{n+1}.
\end{equation}
Early teaching may introduce entire behaviors. Later teaching can focus on the frontier: rare failures, new tools, or missing transitions between familiar blocks. Each attempt reveals what the robot can do and where further guidance would help. This mutual refinement is the educational principle behind the architecture and the practical basis for the prediction that teaching demand should fall as reusable experience accumulates.

\section{From Zero-Shot Agent Control to Teach and Grow}
\label{app:evolution}

\subsection{The Minimal Agent Route}

The project began by asking a multimodal Agent to plan directly from robot-view images. It understood the goal, but understanding the task was not enough to execute it: depth, collision, contact, and high-rate control still demanded robot-native competence.

\subsection{Tool Use and Agentic Visual Feedback}

We then gave the Agent perception, grasping, planning, and control tools, and returned a fresh observation after each short execution. It could now change its route when the world disagreed with the plan. This made zero-shot agent control technically feasible, but learning a physical task could still require many trials and model calls.

Structured memory made those trials useful beyond the current episode: successes supplied strategies, while failures supplied conditions and repairs. The update changed explicit memory rather than model parameters.

\subsection{Why Few-Shot Teaching Is the Practical Default}

Unrestricted exploration is slow and can be unsafe on a physical robot. Few-shot teaching provides a successful starting structure, so autonomous behavior can focus on the variation, correction, and recovery that the demonstrations did not cover. This became the default Teach-and-Grow route.

Multiple sequential model calls make the current Agent more expensive to run than feed-forward policy inference. This motivates the proposed slow-teacher/fast-student design: use the Agent to acquire unfamiliar behavior and produce structured trajectories, then train fast policies for mature tasks. Improvements in model efficiency and caching may also reduce inference cost.

\section{Skill Blocks and Atomic Skills}
\label{app:skillblocks}

\subsection{Why the Term Skill Block}

Atomic skills are an established and useful concept. The word \emph{atomic}, however, often suggests the smallest indivisible controller, action primitive, or short policy. Our decomposition operates at a broader semantic level. A block is one decision-level building element of a task and may contain a substantial internal feedback loop. The block metaphor also fits the scaling argument: new tasks are built by recombining old blocks and learning only the missing pieces.

A short gripper close can be a block when it is independently meaningful. So can a longer \emph{acquire the requested container} behavior that detects, approaches, closes, lifts, and checks retention. The granularity is determined by semantic closure and reuse, not by duration or number of motor commands.

\subsection{Taxonomy}

Skill Blocks can be organized by the kind of effect they produce:
\begin{itemize}
    \item \textbf{acquisition blocks}: establish stable possession or contact;
    \item \textbf{transport blocks}: preserve a state while changing a spatial relation;
    \item \textbf{release and placement blocks}: establish support, containment, alignment, or insertion;
    \item \textbf{articulation and activation blocks}: open, close, rotate, press, toggle, or actuate;
    \item \textbf{observation blocks}: gather evidence needed by later decisions;
    \item \textbf{recovery blocks}: restore a precondition after a diagnosed failure.
\end{itemize}
These families organize retrieval by intended effect. Each block can use a learned policy, planner, servo, or human intervention suited to that effect.

\subsection{A Populated Block from the Experiment}

\begin{table*}[!t]
\caption{A populated acquisition Skill Block induced from three teacher trajectories.}
\label{tab:populated_block}

\centering
\renewcommand{\arraystretch}{1.15}
\setlength{\tabcolsep}{6pt}
\begin{tabular}{@{}p{0.18\textwidth}p{0.72\textwidth}@{}}
\toprule
Field & Stored content \\
\midrule
Subgoal $g$ & Acquire and retain the requested alphabet-soup can. \\
Scope $\mathcal{S}$ & Object-specific in the reviewed LIBERO-Object task; it is not exposed as a generic can skill. \\
Reusable strategy $\rho$ & Bind the current object frame; approach the visible graspable body; align; close with bounded local motion; verify that the object is held; lift while preserving retention. \\
Grounding $\gamma_i(o_t)$ & Estimate current object identity, frame, confidence, obstacles, reachability, grasp candidates, and local motion parameters from fresh observations. \\
Executors $\Pi_i$ & Current-scene grasp selection, servo acquisition, and a newly planned collision-aware transition. \\
Outcome test $v_i$ & Pass only when fresh observations and action feedback verify object retention and displacement. \\
Recovery $\mathcal{R}$ & Reobserve, adjust local parameters, invoke an allowed grasp repair, choose an alternative block, recompose the task, or abort. \\
Excluded & Demonstration world coordinates, pixels, old path, joint trajectory, and low-level action replay. \\
\bottomrule
\end{tabular}
\end{table*}

The second induced block represents \emph{release the held object through the visible opening}. It retains the relation and release strategy; the opening frame, release band, held-object offset, path, and controller commands are computed in the current scene.

\section{Detailed Agent-Led Skill Induction}
\label{app:induction}

\subsection{Source Normalization}

Teaching is first converted into observable event sequences. Robot demonstrations provide synchronized observations and actions; human video provides visible state changes and object relations; written instructions provide intent and order. The source type stays attached because each source supports a different part of a block.

\subsection{Semantic Decomposition}

The Agent splits each trajectory at meaningful state changes using task language, object relations, gripper state, and motion. A segment receives the semantic signature
\begin{equation}
    \sigma(s)=\langle g_s,u_s,a_s,\ell_s,p_s,q_s\rangle,
\end{equation}
where $g_s$ is the subgoal, $u_s$ the manipulated entity, $a_s$ its affordance, $\ell_s$ the target relation, $p_s$ the preceding effect, and $q_s$ the ordering context.

\subsection{Cross-Demonstration and Cross-Embodiment Alignment}

Alignment follows semantic effect rather than motor coordinates. \emph{Acquire the requested container} can preserve the object role and the result \emph{object is stably held} whether a robot uses a parallel gripper, suction, or a dexterous hand. The receiving robot supplies its own executor, grounding, and local verification. This defines a possible route to cross-embodiment reuse. The present study uses one embodiment.

The alignment procedure follows four steps:
\begin{enumerate}
    \item normalize each source into observable events while keeping its type;
    \item align segments that achieve the same effect and have compatible entities or relations;
    \item separate shared task structure from pose, path, timing, and embodiment-specific control;
    \item validate the resulting block on fresh cases and narrow its scope when needed.
\end{enumerate}
Variable-length demonstrations therefore need not have the same number of raw segments. They must agree on the semantic effects used for the final task composition.

\subsection{Invariant Synthesis}

For each aligned group, the Agent separates:
\begin{itemize}
    \item \emph{semantic invariants}: subgoal, object role, relation, and expected effect;
    \item \emph{behavioral invariants}: approach family, contact region, ordering, and permitted executor family;
    \item \emph{instance variables}: pose, pixels, point cloud, grasp, path, timing, and robot configuration.
\end{itemize}
The first two categories define reusable strategy; instance variables are recomputed at runtime. A segment that still depends on the original coordinates remains an episode rather than a block.

\subsection{Scope Selection}

Scope expands only as evidence accumulates. Repeated success across object poses justifies pose invariance, and success across several object instances supports object-level abstraction. A common affordance across categories may justify affordance-level reuse. Contradictory demonstrations call for alternative blocks or narrower conditions, and a single unusual instrument stays object-specific until more evidence arrives.

\subsection{Validation and Library Update}

A candidate block is tested on separated cases before normal retrieval. The test asks whether it produces the intended effect from the current scene, preserves matched old behavior, and stays within its claimed scope. A weak block is narrowed or repaired; a validated one enters a new library version.

\section{Agentic Execution, Memory, and Failure Learning}
\label{app:agent}

\subsection{The Agent's Task-Level Decisions}

At task time, the agent decides:
\begin{itemize}
    \item which observations and modalities are needed;
    \item which Skill Blocks are relevant;
    \item how they should be ordered or conditioned;
    \item which backend should instantiate a block;
    \item whether an observed effect is sufficient to continue;
    \item whether to retry, substitute, recompose, request teaching, or stop;
    \item which parts of an execution are valuable for future learning.
\end{itemize}
The agent makes these decisions at meaningful state transitions, while the block's executor handles continuous control.

\subsection{Success and Failure as Memory}

A successful trace can establish a new block, broaden evidence, or confirm a composition. A failure can be more informative than another success. It may reveal that a visual condition is ambiguous, a grasp family is unsuitable, a planner bridge fails under clutter, or a block's outcome test is too weak. The memory update records a structured explanation and links it to observations and outcomes.

Because the memory is explicit, humans can inspect and edit it. An operator can correct an overgeneralized rule, forbid a recovery, or mark a tool version incompatible. This editability is difficult to achieve when every experience is absorbed only into distributed parameters.

\subsection{Four Levels of Retention}

Persistent storage avoids direct parametric overwriting, yet behavior can still degrade. We distinguish:
\begin{enumerate}
    \item \textbf{storage retention}: the block remains present;
    \item \textbf{retrieval retention}: the correct block is still selected;
    \item \textbf{grounding retention}: its strategy still binds to current sensing and embodiment;
    \item \textbf{behavioral retention}: execution success is maintained.
\end{enumerate}
A longitudinal evaluation measures all four to identify where a retained behavior stops being useful.

\section{Scaling Laws, Cost, and the Teach-and-Grow Hypothesis}
\label{app:scaling}

\subsection{What a Scaling Law Means}

The influential language-model scaling laws did more than state that larger models perform better. They identified explicit resources such as model parameters, training tokens, and compute, then showed that held-out loss follows approximately predictable power-law trends over broad ranges~\cite{kaplan2020scaling}. Chinchilla further showed that the allocation of a fixed compute budget matters: model size and data should grow together near the compute-optimal frontier~\cite{hoffmann2022training}. A useful scaling law needs four things: a measurable resource, an outcome that matters, a stable relationship between them, and a prediction that holds at scales not used to fit the curve.

For robots, the scaling relation must specify which experience counts, what remains fixed, what outcome is predicted, and where the relationship should saturate or fail.

\subsection{What Existing Robot Scaling Laws Measure}

Robot learning already exhibits several valuable scaling regularities. A broad meta-analysis finds power-law-like improvement with data, model size, and compute across many robot-learning studies~\cite{sartor2024neuralscaling}. Large imitation-learning experiments report approximately power-law gains with demonstrations, environments, and object diversity, with diversity often more valuable than repeatedly sampling the same condition~\cite{lin2024datascaling}. Factored scaling curves estimate which environmental factors most deserve additional data~\cite{zha2025factoredscaling}; embodiment scaling studies ask how performance changes as the diversity of bodies grows~\cite{ai2025embodimentscaling}; and agent/world-model work studies pretraining resources and architecture-dependent scaling~\cite{pearce2024agentscaling}. A parallel line spends more computation at inference through sampling, verification, or world-action evaluation~\cite{zhu2025testtimescaling,kwok2025robomonkey,zhao2026wamtesttime}.

These laws answer important questions, but they mostly concern one of three regimes: how a policy improves before deployment as offline resources grow; how a family of policies changes with training diversity; or how one episode improves as test-time computation increases. They do not directly describe a deployed robot that repeatedly turns new experience into persistent, separately reusable competence and then faces another task with a different acquisition burden.

\subsection{Why Performance Scaling and Cost Scaling Diverge}

A benign performance curve can hide a steep data-production curve. Suppose a manipulation domain is broadened along $F$ interacting factors, each with roughly $r$ regimes. Dense coverage scales as $r^F$ in the idealized Cartesian case of Eq.~\ref{eq:dense_coverage}. A learner may generalize between cells and need far fewer demonstrations, but newly discovered interactions still require fresh robot operation. Near a tolerance limit, $\log N_{\mathrm{demo}}\propto1/(P-c)$ with $P>c$ predicts sharply increasing demand as $P\downarrow c$~\cite{xu2026precision}.

For an end-to-end route, a useful cost decomposition is
\begin{equation}
\begin{aligned}
C_{\mathrm{E2E}}(K)={}&C_{\mathrm{collect}}(N_{\mathrm{cover}}(K))
+C_{\mathrm{train}}(N_{\mathrm{cover}}(K))\\
&+C_{\mathrm{regress}}(\theta_{\mathrm{new}}(K),\mathcal{T}_{\mathrm{old}}),
\end{aligned}
\label{eq:e2e_cost}
\end{equation}
Here $K$ counts matched semantic capability additions, $N_{\mathrm{cover}}(K)$ is the interaction needed to cover them, $\theta_{\mathrm{new}}(K)$ is the resulting updated policy, and $\mathcal{T}_{\mathrm{old}}$ is the earlier task set. All cost terms in this section use one preregistered scalar unit, so collection, compute, human time, validation, and regression are converted before they are added. Equation~\ref{eq:e2e_cost} identifies the regime in which dense joint coverage, strict tolerance, and repeated global regression make the next useful increment increasingly costly.

Teach-and-Grow is designed around a different cost structure. In TGL, each counted semantic capability addition is one admitted reusable Skill Block. For block $k$, define
\begin{equation}
\begin{aligned}
\kappa_k={}&\kappa_k^{\mathrm{teach}}+\kappa_k^{\mathrm{ground}}
+\kappa_k^{\mathrm{validate}}+\kappa_k^{\mathrm{link}},\\
C_{\mathrm{TGL}}(K)={}&C_0+\sum_{k=1}^{K}\kappa_k
+C_{\mathrm{retrieve}}^{\mathrm{cum}}(K).
\end{aligned}
\label{eq:tgl_cost}
\end{equation}
Here $C_0$ is the fixed infrastructure cost, and $C_{\mathrm{retrieve}}^{\mathrm{cum}}(K)$ is the retrieval cost accumulated while admitting the first $K$ semantic capability additions, each realized by one block in TGL. If block scope and validation stay local, each $\kappa_k$ remains bounded. With hierarchical retrieval satisfying $C_{\mathrm{retrieve}}^{\mathrm{cum}}(K)=O(K)$, cumulative cost is at most linear in $K$. The robot then pays mainly for the missing behavior and its local interfaces. All-to-all compatibility testing would instead make linking quadratic; block contracts and scopes are what keep the update local.

\begin{figure*}[!t]
\centering
\includegraphics[width=\textwidth]{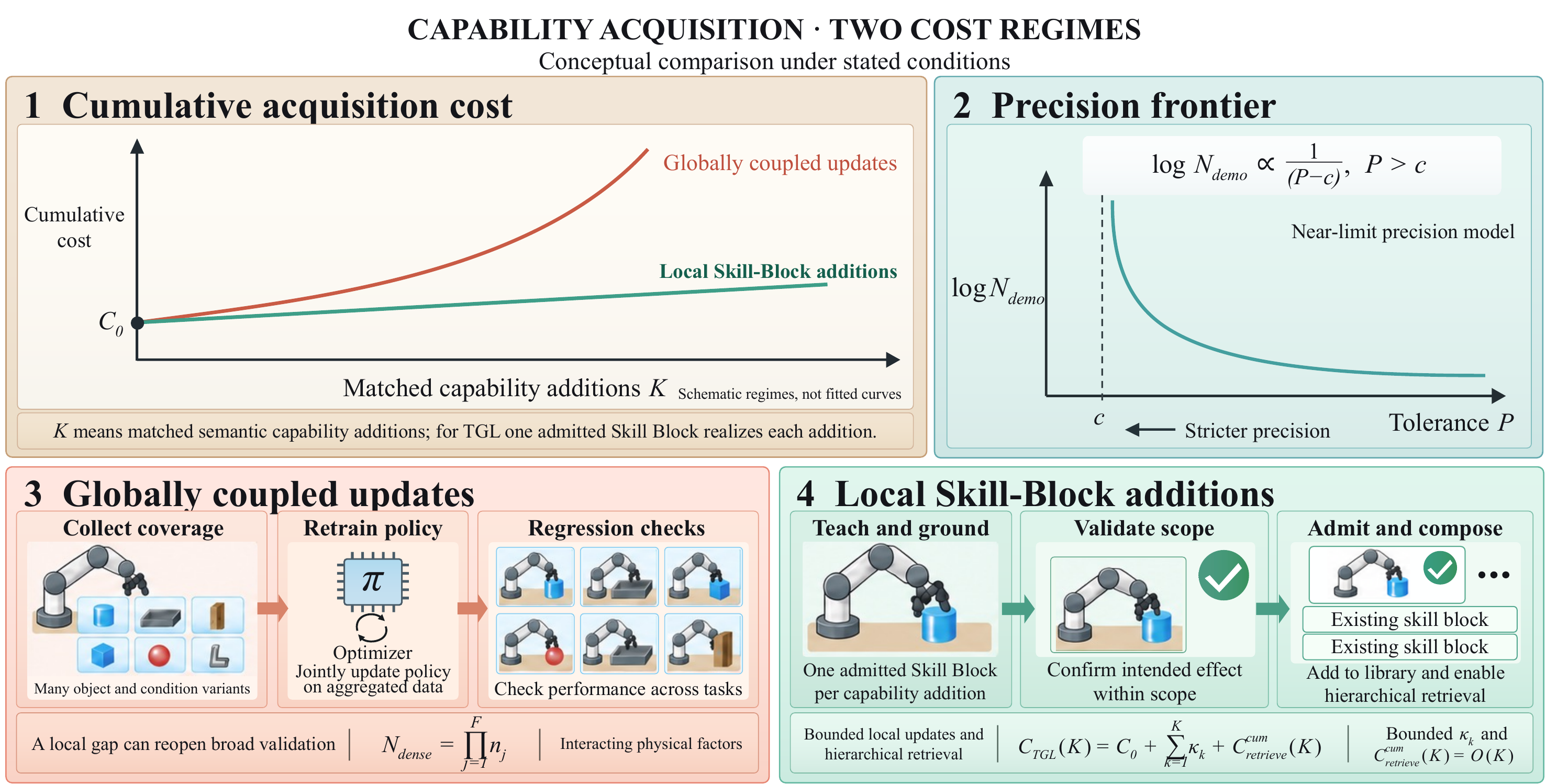}
\caption{Conceptual acquisition-cost regimes. When each addition reopens joint coverage and regression, globally coupled updates can grow convexly with capability breadth. Local Skill-Block additions can remain near-additive when interfaces stay bounded and retrieval is at most linear. Inset: for target tolerance $P>c$, the cited precision law diverges as $P\downarrow c$.}
\label{fig:cost_scaling}

\end{figure*}

Fig.~\ref{fig:cost_scaling} contrasts the two acquisition regimes as a conceptual prediction: broadly coupled updates grow convexly, while local additions can remain near-additive.

\subsection{The New Scaling Resource: Effective Reusable Experience}

The core Teach-and-Grow law concerns benefit rather than storage volume. Equation~\ref{eq:effective_experience} gives each stored experience a score in $[0,1]$ based on five practical factors: evidence reliability, added coverage, retrievability, grounding validity, and compatibility with admitted blocks. The factors and aggregation rule are fixed before the scaling study, and future-task outcomes are hidden from the score.

With the foundation agent, tool versions, test-time budget, future-task distribution, and scoring rule fixed, Eq.~\ref{eq:intro_scaling} predicts that future-task error and teaching demand decay toward irreducible floors as power laws in $X_n$. Equation~\ref{eq:tgl_cost} adds the cost-side prediction: explicit capability can grow close to additively when updates remain local. The resource grows after deployment, and one task can make the next task easier without retraining the full policy.

\subsection{A Longitudinal Test}

A scaling study freezes the foundation agent, tools, inference budget, success criteria, future-task distribution, and scoring rule while the robot learns tasks in sequence. Each library checkpoint receives an $X_n$ value before the next outcomes are observed. The evaluation mixes old tasks, new compositions, new instances of known affordances, and tasks that require a genuinely new behavior. These groups show whether progress comes from remembering an episode, recombining a known skill, broadening its scope, or acquiring a missing block.

At every checkpoint, the primary variables are future-task error $\mathcal{E}_{\mathrm{future}}(X_n)$ and teacher intervention time $D_{\mathrm{teach}}(X_n)$. The latter runs until the fixed success rule or the preregistered teaching-budget cap; budget-exhausted acquisitions remain in the analysis at that cap. Retention, reuse, harmful retrieval, latency, and cumulative acquisition cost explain why a checkpoint departs from the main curve. Raw episodic memory, a fixed tool pipeline, sequential VLA adaptation, and an Agent--VLA hybrid provide matched points of comparison. The power law is fitted on early checkpoints and tested on larger held-out checkpoints alongside standard alternative curves. Total cost includes teaching, robot interaction, inference, verification, storage, retraining, and regression. The prediction is useful only if it reaches unseen checkpoints and reveals where improvement saturates.

\section{A Whole-System View of the Robot Learning Ecosystem}
\label{app:ecosystem}

\begin{figure*}[!t]
\centering
\includegraphics[width=\textwidth]{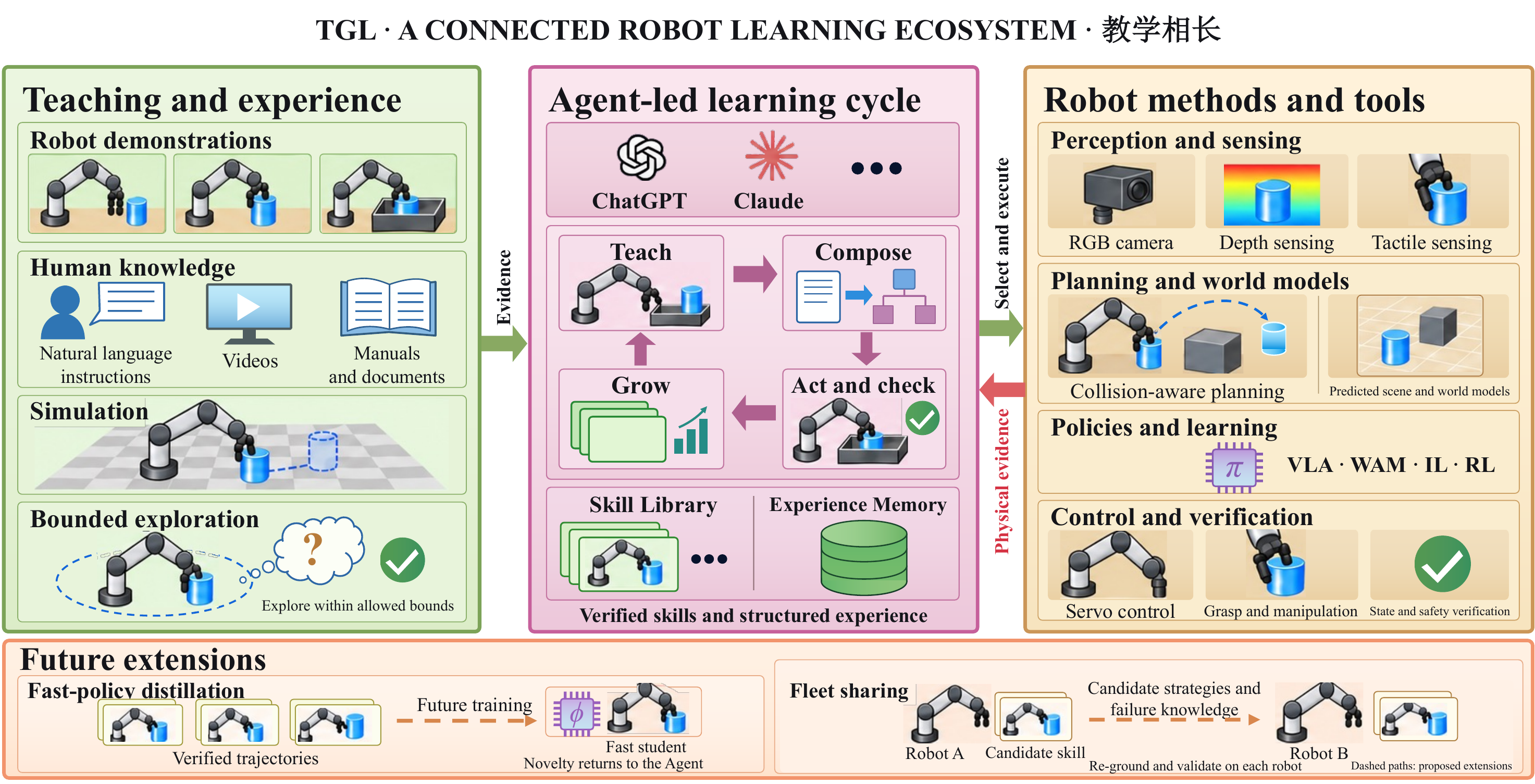}
\caption{Whole-system ecology of Teach-and-Grow Learning. TGL provides a semantic routing and verification layer through which human knowledge, simulation, learned policies, learning methods, perception, planning, control, and safety mechanisms contribute evidence or executable change. The Agent composes these specialized roles around a common learning cycle; verified capability and experience persist separately for later tasks. Dashed paths denote future distillation and fleet sharing.}
\label{fig:ecosystem_whole}

\end{figure*}

\subsection{Sources of Teaching and Experience}

The outer network in Fig.~\ref{fig:ecosystem_whole} accepts more than robot demonstrations. Teleoperation and robot trajectories provide the richest embodiment-aligned evidence and can seed both blocks and fast policies~\cite{khazatsky2024droid}. Simulation, procedural generation, and sim-to-real pipelines can produce controlled variations, failures, and rare conditions before physical deployment~\cite{mandlekar2023mimicgen,nasiriany2024robocasa}. Human demonstration and egocentric video provide inexpensive semantic and interaction structure, although embodiment conversion must recover robot-executable geometry and action~\cite{xu2023xskill,humanego2026}. Manuals, natural-language instructions, diagrams, and CAD can provide ordering, constraints, affordances, and object-specific procedures~\cite{tie2025manual2skill}. Autonomous exploration contributes the complementary evidence that teachers rarely provide: failures, repairs, and unexpected but successful alternatives.

These sources differ in what they can justify. A manual may reveal that a latch must be released before a panel moves, but it cannot by itself certify a grasp controller. Human video may reveal contact sequence and object relation while leaving robot kinematics unresolved. Simulation may offer exact state and broad variation yet still require real-world calibration. TGL preserves source type and uncertainty so that semantic structure can transfer while physical claims are validated by an appropriate executor and observation channel.

\subsection{Roles of Learned Policies and Classical Methods}

A VLA, WAM, diffusion policy, or task-specific policy can occupy three roles in the architecture. It may implement a Skill Block, execute a mature composition as a fast path, or serve as a future student of verified agentic trajectories. Imitation learning and reinforcement learning can build or improve these executors. The surrounding Teach-and-Grow system selects an applicable policy, supplies the evidence it needs, checks its effect, and chooses the next step when execution leaves its supported range.

Classical robotics is equally native to the system. Detection, segmentation, RGB-D geometry, grasp synthesis, motion planning, impedance control, visual servoing, and tactile feedback can each provide evidence or realize a block. Traditional controllers are often the most reliable backend for a well-understood local effect. World models can predict consequences or rank alternatives; test-time verifiers can reject weak candidates; safety filters and humans can constrain the action set. The top-level task is no longer forced into a universal perception--planning--control pipeline. Each block assembles the physical loop appropriate to its own semantic effect.

\subsection{Coordinating Models and Tools through Skill Blocks}

The common interface is a semantic contract: what effect a block is intended to establish, where it applies, what current evidence it requires, which backends can realize it, how success is observed, and what recovery is available. The agent uses those contracts to connect methods that were previously evaluated as separate systems. A visual servo can follow a VLA proposal; a planner can bridge two learned policies; a world model can screen candidate blocks; a human can teach the one transition that simulation failed to discover. The contract allows both classical and learned components to contribute evidence or execute an action within the same block.

Integration remains local: a new module needs its own adapter, calibration, and validation, while unrelated blocks remain unchanged. A tactile sensor can strengthen the blocks that need contact, and a new VLA can replace one executor without erasing a proven visual-servo block. The system can adopt stronger components while preserving the learning cycle around them.

\subsection{Verified Data Generation and Distillation}

A slow agentic learner produces richer supervision than a raw successful trajectory. Its records can include semantic boundaries, selected blocks, current bindings, outcomes, verifier decisions, failure classes, recoveries, and final task success. These signals can train a student to execute one block, choose among blocks, or directly map observations to actions for a mature task family. A practical robot can alternate between acquisition and compression:
\begin{equation}
\begin{aligned}
\text{novel task}&\rightarrow\text{few-shot agent solution}
\rightarrow\text{verified experience}\\
&\rightarrow\text{distilled fast student}
\rightarrow\text{routine deployment}.
\end{aligned}
\end{equation}
Uncertainty, distribution shift, or failure returns control to the agentic route, which expands both the library and the next student dataset. The slow system learns at the frontier; the fast system serves conditions that have become familiar.

\subsection{Personal, Fleet, and Ecosystem Growth}

At the personal level, one robot accumulates skills adapted to its own tools, calibration, and environment. At the fleet level, robots can share candidate semantic strategies, failure signatures, outcome tests, and distilled students; each receiving robot re-grounds and revalidates them with its own embodiment. At the ecosystem level, third parties may publish interaction blocks, specialist perception modules, VLA executors, recovery methods, embodiment adapters, simulators, and validation suites.

Teach-and-Grow connects these contributions within one learning process. Foundation models supply priors; demonstrations, simulation, and human knowledge supply teaching; learned and classical methods supply physical control. The Agent coordinates them, while persistent skills and memory retain useful behavior and context. Future distillation can convert mature behavior into faster execution.

\section{Experimental Details}
\label{app:experiments}

\subsection{Agent Configuration and Trace Provenance}

The benchmark implementation uses OpenAI GPT-6 Astra~\cite{openai2026gpt6astra}, whose official model identifier is \texttt{gpt-6-astra}, with Codex as the agent interface. The two earlier online traces analyzed below used \texttt{gpt-5.6-sol}, seed 0, and structured decisions. In those traces, the Agent received the task, current agent-view and wrist evidence, registered Skill Blocks, allowed tool routes, and verifier or planner evidence. Each trace allowed at most one evidence refresh and two replans; no human changed an in-episode decision.

The two representative successful traces come from a frozen LIBERO-GOAL archive. Together they show the two decisions studied here: rebuilding the route for bowl-on-plate and gathering a new observation for drawer opening.

\subsection{Visual Demonstration Decomposition}

The visual study uses ten successful demonstrations from two related LIBERO-Object condiment-transfer tasks, five per task. Sampled agent-view and wrist images are processed without reward, task-success labels, hidden simulator object state, demonstration actions, or reference boundaries as production inputs. Independent annotations are used only after prediction. The corpus contains 40 predicted and 40 reference stages.

Ordered role/type accuracy is 1.000. Boundary F1 is 0.100 at exact matching, 0.633 within one sampled frame, and 0.900 within two sampled frames. Observable tests verify ten acquisition and ten release effects, producing 20/20 effect confirmations.

\subsection{Task-Specific Skill Learning}

Three teacher trajectories from states 0--2 produce an object-specific acquisition block and a relation-aware release block. On separated states 3--5, the pair succeeds on 3/3 evaluation initial states; the existing six-block route also succeeds on 3/3. After scope checks, the two blocks enter an eight-block library, survive save-and-reload, and again succeed on 3/3. On farther states 6--8, the learned route stops at the missing semantic effect, showing that its scope is enforced during execution.

\subsection{Related-Task Fixed-Executor Pilot}

The pilot compares six-block and eight-block libraries on two related LIBERO-Object tasks. Executor, runtime, seed 0, success criteria, and the three-attempt budget stay fixed; only the learned acquisition and release blocks are added. Library contents and scopes are frozen before evaluation. Onboarding uses states 12--14, and both libraries are evaluated on the same disjoint states 15--17.

With unsuccessful acquisitions retained at the three-attempt cap, the observed median is 2.5 attempts with six blocks and 1.5 with eight. Evaluation success is 0/6 versus 4/6; the corresponding 95\% Wilson intervals are $[0.00,0.39]$ and $[0.30,0.90]$. Both libraries use the same two tasks, seed, initial states, and execution budget.

\subsection{Failure-Localization Cohort}

A separate eight-attempt acquisition cohort records 0/8 task success and localizes the remaining failures: two occur before motion planning, four arise from path consistency or calibration, and two from gripper closure. These categories identify where the feedback loop should improve next.

\section{Current System}
\label{app:implementation}

The GPT-6 Astra/Codex implementation selects blocks and tools, checks outcomes, and revises plans using task instructions and current visual and execution evidence. Detection, segmentation, RGB-D, and robot state ground actions. Contact-GraspNet and MPLib supply grasps and collision-aware motion; controllers execute actions, and observations verify effects. The Skill Library and Experience Memory retain behaviors, scopes, compositions, and context.

\end{document}